\documentclass[conference]{IEEEtran}
\IEEEoverridecommandlockouts

\usepackage{cite}
\usepackage{amsmath,amssymb,amsfonts}
\usepackage{algorithmic}
\usepackage{graphicx}
\usepackage{textcomp}
\usepackage{xcolor}

\usepackage{color}
\usepackage{multirow}
\usepackage{colortbl}
\usepackage{subcaption}
\usepackage{tabularx}
\usepackage{array}
\usepackage{booktabs}
\usepackage{float}
\usepackage{rotating}
\usepackage{hyperref}

\def\BibTeX{{\rm B\kern-.05em{\sc i\kern-.025em b}\kern-.08em
    T\kern-.1667em\lower.7ex\hbox{E}\kern-.125emX}}
\begin{document}

\title{End-to-End Probabilistic Geometry-Guided Regression for 6DoF Object Pose Estimation}

\author{
\IEEEauthorblockN{Thomas Pöllabauer}
\IEEEauthorblockA{\textit{Virtual \& Augmented Reality (VRAR)} \\
\textit{Fraunhofer IGD \& TU Darmstadt}\\
Darmstadt, Germany \\
0000-0003-0075-1181}
\and
\IEEEauthorblockN{Jiayin Li}
\IEEEauthorblockA{\textit{VRAR} \\
\textit{Fraunhofer IGD}\\
Darmstadt, Germany \\
0009-0008-7843-6054}
\and
\IEEEauthorblockN{Volker Knauthe}
\IEEEauthorblockA{\textit{Interactive Graphics Systems Group} \\
\textit{TU Darmstadt}\\
Darmstadt, Germany \\
0000-0001-6993-5099}
\and
\IEEEauthorblockN{Sarah Berkei}
\IEEEauthorblockA{\textit{VRAR} \\
\textit{Fraunhofer IGD}\\
Darmstadt, Germany \\
0000-0002-7986-1414}
\and
\IEEEauthorblockN{Arjan Kuijper}
\IEEEauthorblockA{\textit{Interactive Graphics Systems Group} \\
\textit{TU Darmstadt}\\
Darmstadt, Germany \\
0000-0002-6413-0061}
}

\maketitle

\begin{abstract}
     6D object pose estimation is the problem of identifying the position and orientation of an object relative to a chosen coordinate system, which is a core technology for modern XR applications. 
     State-of-the-art 6D object pose estimators directly predict an object pose given an object observation. Due to the ill-posed nature of the pose estimation problem, where multiple different poses can correspond to a single observation, generating additional plausible estimates per observation can be valuable.
     To address this, we reformulate the state-of-the-art algorithm GDRNPP and introduce EPRO-GDR (End-to-End Probabilistic Geometry-Guided Regression). Instead of predicting a single pose per detection, we estimate a probability density distribution of the pose. 
     Using the evaluation procedure defined by the BOP (Benchmark for 6D Object Pose Estimation) Challenge, we test our approach on four of its core datasets and demonstrate superior quantitative results for EPRO-GDR on LM-O, YCB-V, and ITODD. Our probabilistic solution shows that predicting a pose distribution instead of a single pose can improve state-of-the-art single-view pose estimation while providing the additional benefit of being able to sample multiple meaningful pose candidates.
\end{abstract}


%
%
\section{Introduction}
\label{sec:intro}
Detecting objects in 3D space, relative to a camera, is an essential problem for robotics and XR applications. Current state-of-the-art object pose estimators achieve good results predicting a single pose, given a novel object observation. At the same time, the pose estimation task is an ill-posed one, as scene characteristics such as occlusion can drastically reduce estimation accuracy. An even more serious problem is the pose ambiguity problem, where multiple poses may explain a certain observation. We argue that the pose ambiguity problem can be mitigated by predicting a probability density function (pdf) instead of discrete poses.
Towards this end, we enhance and reformulate the state-of-the-art algorithm GDRNPP and introduce EPRO-GDR. Given a single observation, EPRO-GDR is capable of sampling multiple meaningful poses together with an expressive representation of the uncertainty of any sampled pose. While this approach is particularly useful for scene-level optimization, where the poses of all objects in a scene are optimized together, it also enhances the accuracy of individual object pose estimates, as we demonstrate. \\

Our contribution consists of improving the state-of-the-art single-view pose estimator GDRNPP to predict a pdf per detected object and image, instead of directly regressing pose. EPRO-GDR allows sampling multiple relevant pose candidates together with a meaningful measure of uncertainty, which we argue is useful for multi-view scene-level pose optimization. At the same time, it outperforms the baseline GDRNPP on 3 out of 4 BOP core datasets. 

\section{Related Work}
\label{sec:relatedWork}
6D object pose estimation methods can be divided into traditional methods and methods based on deep learning \cite{20.500.11850/583168,zhu2022review}. Recently the state-of-the-art has been dominated by machine learning-based approaches, as can be seen in the results of the representative BOP challenge \cite{hodan2024bop,sundermeyer2023bop}. In our discussion we will focus on these ML-based algorithms.

\subsection{Direct Regression 6D Object Pose Estimation}
A first group of algorithms uses regression, directly predicting the object pose based on a given image \cite{zhu2022review}. Methods of this kind are SSD-6D \cite{kehl2017ssd6d}, PoseCNN \cite{xiang2018posecnn}, BB8 \cite{rad2017bb8}, and RDPN \cite{rdpn}.
SSD-6D follows a similar idea as the predating SSD detector: approaching the detection task by classifying dense candidate boxes and regressing their locations. However, it extends this idea by discretizing the rotation space and transforming the regression problem into a classification task. 
PoseCNN leverages convolutional neural networks (CNN) to extract features from RGB images and predicts the semantic labels, rotation, and translation. The paper notably introduces the widely used YCB-Video (YCB-V) dataset, which serves as a core dataset within the BOP challenge.
BB8 predicts the 6D poses using a CNN, represented as 2D projections of the corners of their 6D bounding boxes. To deal with rotational symmetric objects PoseCNN restricts the range of poses used for training and introduces a classifier to identify the pose range at runtime before estimating it. Additionally, an optional refinement step is employed to improve the accuracy of the predicted poses.
RDPN predicts dense correspondences, specifically the object coordinates per visible pixel. Using conventional object detection they crop objects from the image and adjust the camera intrinsic per crop to fit the observation. Next, RDPN extracts relevant features to obtain an object mask and object coordinates. In a third step, dense correspondences are established and used in the pose predictor for pose estimation. RDPN requires RGB-D input.

\subsection{Perspective-$n$-Point}
\subsubsection{P$n$P-based Pose Estimators}
Another group of algorithms predicts various kinds of keypoints and key image features and solves for the pose using perspective-$n$-point algorithm (P$n$P). Among these are SingleShotPose \cite{tekin2018real}, PVNet \cite{peng2018pvnet}, HybridPose \cite{song2020hybridpose}, CDPN \cite{cdpn}, EPOS \cite{hodan2020epos}, SurfEmb \cite{haugaard2022surfemb}, ZebraPose \cite{su2022zebrapose}, GDR-Net \cite{gdrnet}, and its improved version GDRNPP \cite{liu2022gdrnpp_bop}. 
SingleShotPose extends a YOLO detector to predict the 8 axis-aligned 6D bounding box corners and uses P$n$P to solve for the pose. 
PVNet generates a voting vector for each pixel in the input images. These voting vectors are then aggregated, resulting in the final prediction of key points. Additionally, a mask is predicted to filter out irrelevant pixels, thereby enhancing accuracy. 
HybridPose employs a hybrid intermediate representation to capture various geometric information present in the input image, such as keypoints, edge vectors, and symmetry correspondences. These different intermediate representations can all be predicted using a single neural network. Additionally, a robust regression module is utilized to filter out outliers in the predicted intermediate representations.
CDPN employs a disentangled approach to predict rotation and translation separately for pose estimation. It utilizes a CNN to predict a dense map of 3D coordinates and a mask. The rotation is obtained by solving P$n$P/RANSAC using the predicted coordinates, while the translation is estimated directly from the image. 
In EPOS, objects are represented by compact surface segments. A network with an encoder-decoder architecture is employed to predict the correspondence between densely sampled pixels and these segments. The pose is subsequently determined using a modified P$n$P/RANSAC method called graph-cut RANSAC \cite{barath2018graph}.
SurfEmb learns pixel-wise surface distributions and a mask to establish a correspondence distribution. For each individual point on the object in a 2D image, there is a corresponding area on the surface of the 3D models. The pose hypotheses are obtained using P$n$P/RANSAC and are later evaluated using the surface distributions and masks. The pose hypothesis with the highest score is then refined and optimized based on the 2D-3D correspondence distributions.
ZebraPose employs a discrete descriptor that provides dense representation of the object surface instead of relying on learning dense maps. This descriptor utilizes a hierarchical binary surface encoding as an intermediate representation for 3D coordinates, offering increased robustness against occlusion. The final pose estimation is achieved by solving the P$n$P problem using the Progressive-X method \cite{barath2019progressivex}.
GDR-Net predicts intermediate geometric features, including dense correspondences and surface region attention. Subsequently, the Patch-P$n$P algorithm directly regresses the 6D object pose. This approach makes GDR-Net differentiable, distinguishing it from traditional two-stage pipelines that establish 2D-3D correspondences and then utilize a variant of the P$n$P/RANSAC algorithm. The differentiable nature of Patch-P$n$P (it is a neural network) makes it particularly suitable for tasks that necessitate differentiable poses.
GDRNPP is an enhanced version of GDR-Net (Geometry-Guided Direct Regression Network) that incorporates stronger domain randomization operations for augmentation. Additionally, ResNet-34 \cite{resnet} is replaced with ConvNeXt \cite{liu2022convnet} and instead of only predicting the visible mask, it predicts the amodal mask as well.

\subsubsection{P$n$P Algorithms}
There are many different algorithms to solve the P$n$P problem, among them are Iterative P$n$P, Efficient P$n$P (EP$n$P) \cite{lep2009epnp}, Generalized End-to-End Probabilistic P$n$P (EPro-P$n$P) \cite{chen2023epropnp}, Progressive-X (Prog-X) \cite{barath2019progressivex}, and Patch-P$n$P \cite{gdrnet}. 
Iterative P$n$P refines an initial estimate by minimizing the re-projection error until it falls below a certain threshold. 
EP$n$P achieves its efficiency by expressing pose as a function of four virtual control points, thereby reducing the number of unknowns to be solved. 
EPro-P$n$P is a probabilistic approach and differentiable. It minimizes the Kullback-Leibler divergence between prediction and target pose distribution to learn the intermediate variables: 2D-3D coordinates and corresponding 2D weights.
Prog-X progessively grows the set of points considered to find a solution, allowing to incrementally improve upon the current solution.

\subsection{Template Matching}
Template matching tries to solve for object pose by generating templates (often renderings based on 3D meshes) and matching them with the object appearances found in the target view. This often involves finding key points in both the template and the image and looking for similarities. One of the most prominent methods following this idea is the non-ML algorithm Linemod \cite{hinterstoisser2013model}. 
But the idea is also found in more modern algorithms, such as the zero-shot methods MegaPose \cite{labbe2022megapose} and GigaPose \cite{nguyen2023gigapose}. PFA-Pose (Prspective Flow Aggregation) \cite{hu2022perspective}, after estimating an initial pose using a first network, continues to match this pose with offline-generated templates. However, for the BOP challenge, the rendering is done online, resulting in improved accuracy compared to offline rendering as mentioned in \cite{PFA-BOP}. Comparison between retrieved template and target view in the 2D image is done by computing the displacement field between both. This field represents the distance and direction that each pixel needs to move from the example image to the target image. Displacement field results are combined and transformed into a set of 3D-2D correspondences. The final result is obtained by solving the P$n$P problem using RANSAC/P$n$P. As a result, this method consists of two stages and cannot be trained end-to-end, which we think is a major drawback and important reason to choose to extend GDRNPP over PFA. 

\begin{figure*}[h]
    \centering
    \includegraphics[width=0.8\textwidth]{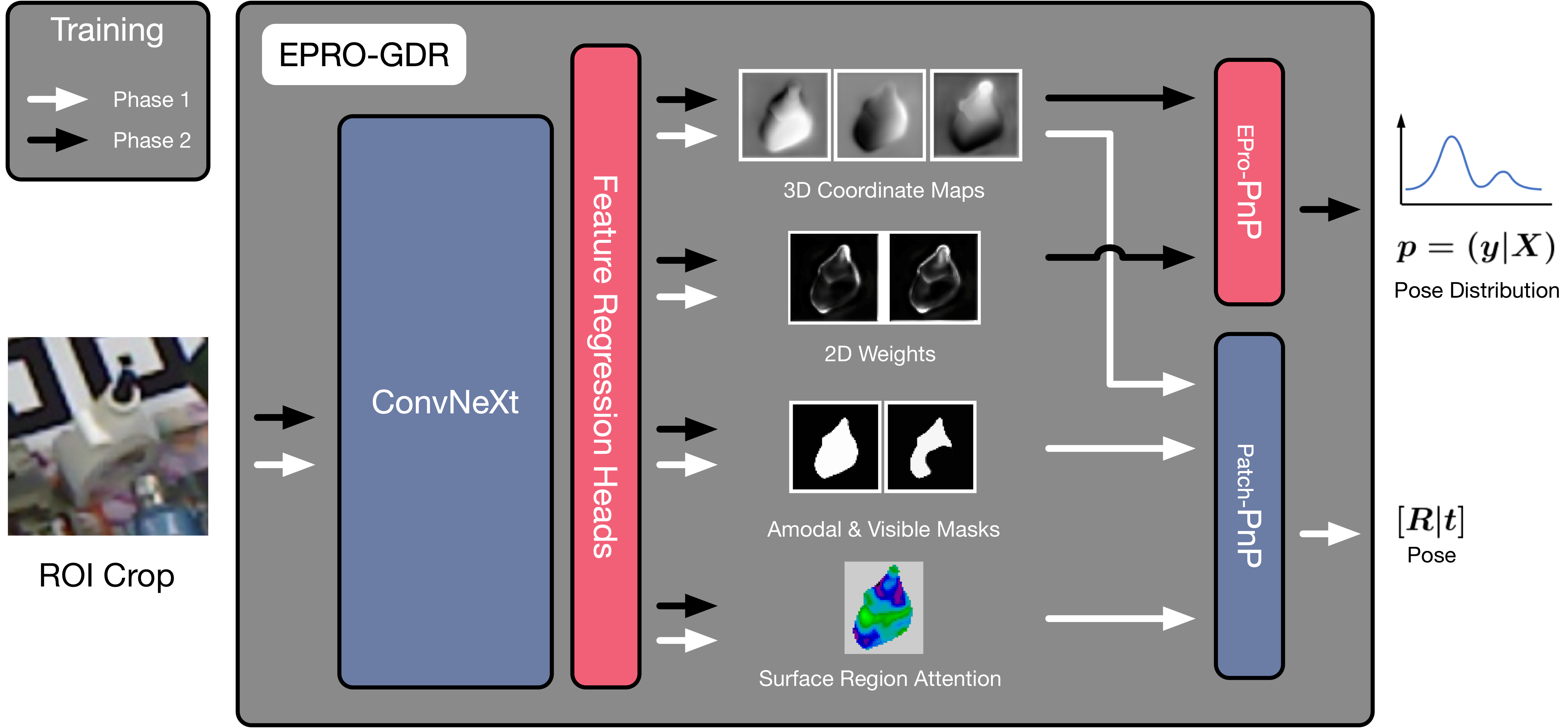}
    \caption{Proposed method with both training phases. We start by training GDRNPP as described in the author's description of their entry to the BOP challenge \cite{liu2022gdrnpp_bop}, predicting 3D coordinate maps, masks, and surface region attention and solving directly for pose using Patch-P$n$P (Phase 1, white arrows). Phase 2 / black arrows: After convergence we replace Patch-P$n$P with EPro-P$n$P and modify the loss functions (details in Section \ref{sec:implementation}), also predict the 2D weights required by EPro-P$n$P, and continue training. After convergence, our model is fully trained. At inference, we predict a distribution instead of a single pose. The data flow is the same as with Phase 2, our second training phase.  }
    \label{fig:method_diagram}
\end{figure*}

\subsection{Pose Refinement}
Pose refinement is the process of starting with a coarse initial pose estimate and using additional (often iterative) steps to (step-wise) increase the final prediction accuracy. 
DeepIM \cite{Li_2019} builds upon PoseCNN and introduces an iterative optimization process to enhance pose estimation accuracy. This process involves iteratively matching between images rendered from the 3D model. The refined pose from each iteration serves as the input for the next iteration. This iterative matching continues until the process converges or reaches the maximum number of steps. CosyPose, as described in \cite{labbé2020cosypose}, combines the DeepIM approach with scene-level refinement. This approach improves both pose estimation and correspondences simultaneously. RePose, introduced in \cite{repose}, achieves faster runtime by replacing repeated forward passes of CNN-based optimization with a fast renderer and a learned 3D texture.
Another important refinement method, Iterative closest point (ICP) \cite{121791}, is commonly used for aligning 3D models. Given two point clouds, the algorithm iteratively adjusts the transformation parameters, including rotation and translation, to minimize the distance between corresponding points in the two point clouds.
Coupled-iterative refinement (CIR) \cite{lipson2022coupled} involves estimating the flow between an input image and a collection of rendered images of a known 3D object. This estimation process generates 2D-3D correspondences, which are subsequently utilized to solve for pose estimation. What sets CIR apart is its coupled iterative approach, where the flow and object pose are updated in a mutually dependent manner. Specifically, the update of the flow is conditioned on the current pose and vice versa.
GDRNPP \cite{liu2022gdrnpp_bop} propose a depth refinement step to improve the accuracy of the translation estimate. This refinement is achieved by comparing the rendered object depth and the observed depth. The depth refinement value is computed as the median of the differences between these depth values for corresponding pixels. The updated translation vector is then obtained by adding a translation adjustment, calculated based on the depth difference and the inverse of the camera intrinsic matrix, to the initial translation estimate.

\section{Proposed Method}

We discuss the algorithm selection for probabilistic pose estimation, discuss our modifications, and overall approach, before detailing our implementation and training details.


\subsection{Algorithm Selection}
We can draw some conclusions from previous work: First, P$n$P-based methods tend to be more likely to achieve SotA performance. Second, refinement plays a critical role to lift a good result to one of the best. For instance CIR increases average recall of GDRNPP-PBR-RGB-MModel by approximately 0.1, as shown in the BOP challenge \cite{sundermeyer2023bop}. Finally, probability-based methods \cite{chen2023epropnp,haugaard2022surfemb,haugaard2024spyropose} modeling pose distributions show the potential to increase robustness. In our opinion, the ability to sample multiple plausible pose candidates per detection, along with their associated probabilities as an indicator of quality, makes them an excellent choice for scene-level optimization.\\

\begin{table}[]
    \centering
    \begin{small}
    \caption{Top 10 entries in the leaderboard of the BOP challenge for the task localization of seen objects \cite{bop_leaderboard}.}
    \begin{tabular}{|c|l|c|c|}
      \hline
        Rank & Method & AR & Time \\
        \hline
        1 & GPose2023 & 85.6 & 2.670 \\
        \hline
        2 & GPose2023-OfficialDetection & 85.1 & 4.575 \\
        \hline
        3 & GPose2023-PBR & 84.4 & 2.686 \\
        \hline
        4 & GDRNPP-PBRReal-RGBD & 83.7 & 6.263 \\
        \hline
        5 & GDRNPP-PBR-RGBD & 82.7 & 6.264 \\
        \hline
        6 & ZebraPose-EffnetB4-refined & 81.3 & 2.577 \\
        \hline
        7 & GDRNPP-PBRReal-RGBD-Fast & 80.5 & 0.228 \\
        \hline
        8 & PFA-Mixpbr-RGBD & 80.0 & 1.193 \\
        \hline
        9 & RDPN & 79.8 & 2.429 \\
        \hline
        10 & GDRNPP-PBRReal-RGBD-OfficialDet. & 79.8 & 6.406 \\
        \hline
    \end{tabular}
    \label{tab:bop_leaderboard}
    \end{small}
\end{table}

To select a suitable pose estimation algorithm to extend we take a look at the BOP leaderboard presented in Table \ref{tab:bop_leaderboard}. We find five different algorithms among the top 10 positions: GPose, GDRNPP, ZebraPose, PFA, and RDPN. GPose is an extension to GDRNPP, though there is no paper nor an implementation published. GDRNPP, ZebraPose, PFA, and RDPN were discussed above in Seciton \ref{sec:relatedWork}. Based on our algorithm discussion and their ranking in the BOP leaderboard, we choose to extend the SotA method GDRNPP for several reasongs: first, for its simple design and high performance in the BOP Challenge. The simple encoder-decoder architecture and the realization of P$n$P in form of a neural network are appealing. Most importantly, because of this design, GDRNPP can be trained in an end-to-end (and fully differentiable) manner and it is possible to directly concatenate any features that may facilitate the resolution of the P$n$P problem and input them into the implicit P$n$P solver, as has been shown in a straightforward stereo vision extension \cite{pollabauer2024extending}. Also, GPose as the overall best performing algorithm being based on GDRNPP shows there is still potential in improving the architecture. GDRNPP's design and use of Patch-P$n$P also makes it a natural candidate to incorporate a probabilistic method for pose distribution prediction, which we argue can benefit (multi-view) scene-level optimization. Based on our research, EPro-P$n$P seems to be a natural fit for our purpose and we decide to integrate EPro-P$n$P into GDRNPP and call our modified algorithm EPRO-GDR. 

\subsection{Implementation Details}
\label{sec:implementation}


We propose to replace GDRNPPs Patch-P$n$P algorithm with EPro-P$n$P \cite{chen2023epropnp} to easily sample multiple pose candidates per image and  object detection. We continue by presenting EPRO-GDR in detail.

\label{sec:eprogdr}

GDRNPP runs a detector (YOLOX \cite{ge2021yolox} in case of the BOP challenge) to detect objects within the 2D image grid, crops the object and feeds the image patch (RoI) to the backbone network ConvNeXt. Inspired by CDPN the extracted feature maps are used to predict 3 different features: visible and amodal object masks $M_{vis}$ and $M_{amodal}$, 3D coordinate maps $x^{3D}$ transformed into dense correspondence maps $M_{2D-3D}$, and surface region attention maps $M_{SRA}$. Compared to CDPN the translation head is removed. The extracted features get handed to the Patch-P$n$P solver consisting of a small CNN with 3 layers leading into 2-layer MLP. 

EPro-P$n$P extracts 3D coordinate maps and 2D-3D correspondence maps to solve the P$n$P problem. Based on the architecture of CDPN, EPro-P$n$P proposes to view the P$n$P problem as a non-linear least squares problem written in equation \ref{eq:non-linear least squares problem} \cite{chen2023epropnp}:
\begin{equation}
\underset{y}{\operatorname{arg\ min}} \frac{1}{2} \sum_{i=1}^N\|\underbrace{w_i^{2 \mathrm{D}} \circ\left(\pi\left(R x_i^{3 \mathrm{D}}+t\right)-x_i^{2 \mathrm{D}}\right)}_{f_i(y) \in \mathbb{R}^2}\|^2
\label{eq:non-linear least squares problem}
\end{equation}
The objective is to estimate a target pose $y$ minimizing the cumulative squared weighted re-projection error. Using the projection function $\pi(\cdot)$, along with an element-wise product denoted by $\circ$, the 2D points in the image are computed based on the predicted pose and the intrinsic properties of the camera and 3D points. Subsequently, the differences between these computed 2D points and the ground truth 2D points are calculated. These differences are then multiplied by the predicted weights of the correspondences, denoted as $f_i(y)$. Instead of solving only for a singular solution EPro-P$n$P suggests to model the prediction as a distribution, accommodating the fact that there are many non-distinguishable solutions of the non-linear least squares problem. Please refer to the base paper \cite{chen2023epropnp} for details on the inner workings of EPro-P$n$P.
Just like GDRNPP, EPro-P$n$P borrows heavily from the design of CDPN and only adds slight modifications to the outputs. Most importantly the translation head is, again, removed. EPro-P$n$P, just like GDRNPP, predicts 3D coordinate maps $x^{3D}$, and additional 2D XY weight maps $w^{2D}$, taken from CDPN, but extending it with spatial Softmax and global scaling. 

Based on these two algorithms we develop our approach: 
First, while EPro-P$n$P originally utilizes a ResNet-34 backbone, we instead decide to re-use GDRNPPs ConvNeXt, saving computation time by sharing the common feature representation among all 4 prediction heads. We predict all 3 features found in GDRNPP, visible and amodal object masks $M_{vis}$ and $M_{amodal}$, 3D coordinate maps $x^{3D}$ transformed into dense correspondence maps $M_{2D-3D}$, and surface region attention maps $M_{SRA}$ and additionally predict the 2D XY weights $w^{2D}$ proposed by CDPN and modified by EPro-P$n$P. 
Second, we find that EPRO-GDR benefits from good initialization and decide to split the training in 2 phases: We utilize the Patch-P$n$P solver and train till convergence, only predicting $M_{vis}$, $M_{amodal}$, $x^{3D}$/$M_{2D-3D}$ and $M_{SRA}$, as depicted in Figure \ref{fig:method_diagram}. We then replace Patch-P$n$P with EPro-P$n$P and continue training, now only using 3D coordinate maps $x^{3D}$, as well as our newly introduced head, predicting 2D weights $w^{2D}$. 
Third, we modify the loss functions in phase 2: Most importantly, we add a new term from EPro-P$n$P, namely Kullback-Leibler loss (KL), measuring the divergence between predicted and target pose distributions. Since we do not require surface region attention for EPro-P$n$P, in phase 2 of our training, we reduce its weight from 1.0 in phase 1 to 0.005. In our experiments this made it behave like regularization, while not interfering with our newly introduced KL loss, which we weight with 0.2. Also, we activate the rotation loss term found in GDRNPP. They set it to 0.0 in their \href{https://github.com/shanice-l/gdrnpp_bop2022/blob/8c6c34b1705008c6798f921dc5609f1e77f045e2/configs/_base_/gdrn_base.py#L138}{config file} 
and do not use it in their \href{https://github.com/shanice-l/gdrnpp_bop2022/blob/8c6c34b1705008c6798f921dc5609f1e77f045e2/configs/gdrn/lmo_pbr/convnext_a6_AugCosyAAEGray_BG05_mlL1_DMask_amodalClipBox_classAware_lmo.py}{training config} 
. It is defined as 
\begin{equation}
    L_{\text{rot.}} = \frac{1}{2} \left(1 - \frac{\text{trace}(\text{m}_{1} \cdot \text{m}_{2}^T) - 1}{2} \right)
\end{equation}
with $m_{1}$ and $m_{2}$ being rotation matrices and $m_{2}^T$ the transpose of $m_{2}$. The equation calculates the angular distance between the predicted and the ground truth rotation, that is, the smallest angle required to align both rotations. We found angular distance loss helpful and set its weight to 1.0. Following this training regimen we achieve the performance as presented in Section \ref{sec:evaluation}. Training details are to be found in Section \ref{sec:training}. \\
At inference, we can predict a pose distribution given only an RGB image. We start by filtering low confidence 2D-3D correspondences to compute the initial pose before refining the pose relying on Levenberg-Marquardt algorithm to converge to our final best pose estimate. If available, we use the additional channel of an RGB-D input to refine our translation vector further using depth refinement, following the methodology of GDRNPP.

\subsection{Training Details}
\label{sec:training}
We train one model per dataset and only use BlenderProc-generated \cite{denninger2020blenderproc} PBR images as provided by the BOP challenge. As mentioned above, the training consists of two phases: In phase 1 we rely on Patch-P$n$P to solve for our target pose and rely on the default parameterization of GDRNPP as used in the BOP challenge, while in phase 2 we replace Patch-P$n$P with EPro-P$n$P and continue training using our modified loss function. We base our code on GDRNPP and use all of its augmentation such as strong randomization and dynamic zoom-in on the object crops (also used by CDPN and EPro-P$n$P). For phase 2 we increase the batch size from 48 to 72, do 400 iterations warm-up to gently introduce the modified loss function and re-weighted loss terms, then keep the learning rate stable before ending the training with cosine annealing. As proposed with GDRNPP we rely on the Ranger optimizer \cite{ranger} and choose a learning rate of 0.0008. Weight decay is set to 0.01 and we train for 40 epochs, using early stopping based on our disjoint validation set.
\section{Evaluation}
\label{sec:evaluation}

\begin{table*}
\centering
\small
\caption[]{Single-View Results on LM-O, YCB-V, ITODD, and T-LESS. Each algorithm is trained per dataset (single model, multiple objects). $AR_{BOP}$ metric following the official BOP approach, higher is better. The best result between GDRNPP and EPRO-GDR in \textbf{bold font}.}
\begin{tabularx}{\textwidth}{c|XXXX|XXXX|XXXX|XXXX}
& \multicolumn{4}{c|}{LM-O} & \multicolumn{4}{c|}{YCB-V} & \multicolumn{4}{c|}{ITODD} & \multicolumn{4}{c}{T-LESS} \\
& MSPD & MSSD & VSD & Mean & MSPD & MSSD & VSD & Mean & MSPD & MSSD & VSD & Mean & MSPD & MSSD & VSD & Mean \\ \hline
SurfEmb \cite{bop2021surfemb} & 0.856 & 0.809 & 0.615 & 0.760 & 0.792 & 0.849 & 0.757 & 0.757 & 0.560 & 0.558 & 0.497 & 0.538 & 0.859 & 0.829 & 0.797 & 0.828 \\
PFA \cite{PFA-BOP} & 0.890 & 0.843 & 0.658 & 0.797 & 0.881 & 0.920 & 0.863 & 0.888 & 0.498 & 0.495 & 0.413 & 0.469 & 0.825 & 0.795 & 0.718 & 0.779 \\
\rowcolor[gray]{0.9} GDRNPP \cite{gdrnpp_bop} & 0.849 & 0.803 & 0.619 & 0.757 & 0.814 & 0.876 & 0.761 & 0.817 & 0.370 & 0.397 & 0.302 & 0.356 & \textbf{0.903} & \textbf{0.871} & \textbf{0.793} & \textbf{0.856} \\
\rowcolor[gray]{0.9} EPRO-GDR (Ours) & \textbf{0.879} & \textbf{0.835} & \textbf{0.645} & \textbf{0.786} & \textbf{0.832} & \textbf{0.894} & \textbf{0.807} & \textbf{0.844} & \textbf{0.447} & \textbf{0.433} & \textbf{0.358} & \textbf{0.412} & 0.811 & 0.769 & 0.715 & 0.765 \\
\end{tabularx}
\label{tab:quantitative1}
\end{table*}

For our quantitative analysis, the standardized tasks and tests in the BOP challenge allow for a fair comparison with the wide landscape of pose estimation algorithms and we comply with their evaluation methods using the BOP toolkit. In addition to our base algorithm GDRNPP and for comparison with pose estimators following different paradigms, we utilize two additional algorithms, probability-based SurfEmb \cite{haugaard2022surfemb} and template-based PFA \cite{hu2022perspective} as comparison. Both perform at the top of the class in their respective paradigm as shown in the BOP leaderboard. We evaluate on a representative subset of the 7 BOP core datasets, namely LM-O \cite{lmo}, YCB-V \cite{xiang2018posecnn}, T-Less \cite{tless}, and ITODD \cite{itodd}. LM-O presents 8 objects under heavy occlusion, placed in cluttered scenes and varying lighting conditions. Among the objects are some with very little texture. YCB-Video has 21 household and food items, such as cracker boxes, and cans and is mostly strongly textured. ITODD has gray scale recordings featuring 28 real-world objects found in industrial use cases including objects with metallic surfaces. Finally, T-Less comprises 30 objects found in the context of electric installations. The real-world objects have little texture and the test images feature heavy occlusions, a high number of object instances per image, and varying lighting conditions. We report results per dataset in Table \ref{tab:quantitative1} showing the 3 metrics used in the BOP challenge, maximum symmetry-aware projection distance (MSPD), maximum symmetry-aware surface distance (MSSD), and visible surface discrepancy (VSD). VSD quantifies the difference between the visible surfaces of two 3D objects, while MSSD measures the maximum distance between corresponding points on surfaces, taking into account the inherent symmetry of the models. MSPD calculates the maximum distance between corresponding points using the projection, measuring the perceivable deviation. A pose is considered correct if the deviation according to the metric falls below a given threshold. The overall score in the BOP challenge is defined as the average of these 3 metrics. In addition, we report the widely used ADD-S 0.1 metric, though we find a strong correlation with the BOP metrics and think them more meaningful when trying to evaluate the expected performance for a given use case and use $AR_{BOP}$ for discussing performance on the troublesome T-Less dataset. The ADD metric measures the average distance between model points transformed by the estimated pose and those transformed by the ground truth pose. Recall is determined by a correctness threshold, typically set at 10\% of the object's diameter (ADD 0.1). To account for object symmetries (ADD-S), evaluation can compute the distance between each transformed model point and the nearest point on the ground truth model, thus accommodating equivalent transformations to yield identical scores.
To discuss the effect of our method on individual objects in T-Less, we show per object results in Table \ref{tab:quantitative3} using the $AR_{BOP}$ metrics. 

\subsection{Quantitative Results}

We can see a noticeable outperformance on 3 out of 4 sets, LM-O, YCB-V, and ITODD and perform especially well on the notoriously difficult ITODD dataset with an increase of more than 10\% in the $AR_{BOP}$ metrics. We will now take a closer look at the per-dataset results. \\ 
\textbf{LM-O.} With LM-O we increase the $AR_{BOP}$ score by 2.9\%, from 0.757 to 0.786, and the ADD-S 0.1 score from 85.72 to 88.65 as presented in Table \ref{tab:lmo}. Considering the $AR_{BOP}$ scores, we outperform GDRNPP on 8 out of 8 objects. The biggest (relative as well as absolute) gain is to be found on the symmetric object eggbox, followed by the texture-less cat. Like stated above, LM-O shows heavily occluded objects with little texture in a cluttered environment under high variation in lighting, and EPRO-GDR shows a clear outperformance compared to its predecessor GDRNPP. 

\begin{table}[H]
\centering
\caption{Results with per object results on LM-O. ADD-S 0.1 metric, higher is better. The best results in \textbf{bold font}. EPRO-GDR achieves the best overall performance among all methods.}
\begin{tabular}{lcccc}
\toprule
\multirow{1}{*}{Objects} & \multicolumn{1}{c}{SurfEmb\cite{bop2021surfemb}} & \multicolumn{1}{c}{PFA\cite{PFA-BOP}} & \multicolumn{1}{c}{GDRNPP\cite{gdrnpp_bop}} & \multicolumn{1}{c}{Ours} \\
\midrule
ape & 70.86 & 75.43 & \textbf{77.71} & 76.57 \\
can & 95.98 & \textbf{98.99} & 97.49 & 98.49 \\
cat & 76.02 & 86.55 & 84.21 & \textbf{88.89} \\
driller & 95.50 & \textbf{96.50} & 95.50 & \textbf{96.50} \\
duck & 79.44 & 81.11 & 80.56 & \textbf{82.78} \\
eggbox & \textbf{80.56} & 72.78 & 60.56 & 74.44 \\
glue & 92.86 & \textbf{95.71} & \textbf{95.71} & 95.00 \\
holepuncher & 93.50 & \textbf{98.00} & 94.00 & 96.50 \\
\hline
Mean & 85.59 & 88.13 & 85.72 & \textbf{88.65} \\
\bottomrule
\end{tabular}
\label{tab:lmo}
\end{table}

\begin{table}[h!]
\centering
\scriptsize
\caption{Results with per object results on YCB-V. ADD-S 0.1 metric, higher is better. The best results in \textbf{bold font}. EPRO-GDR achieves the best overall performance.}
\begin{tabular}{lcccc}

\toprule
\multirow{1}{*}{Objects} & \multicolumn{1}{c}{SurfEmb\cite{bop2021surfemb}} & \multicolumn{1}{c}{PFA\cite{PFA-BOP}} & \multicolumn{1}{c}{GDRNPP\cite{gdrnpp_bop}} & \multicolumn{1}{c}{Ours} \\
\midrule

002\_master\_chef\_can & \textbf{100.00} & 96.67 & \textbf{100.00} & \textbf{100.00} \\
003\_cracker\_box & \textbf{100.00} & 92.89 & 60.44 & 73.33 \\
004\_sugar\_box & \textbf{100.00} & \textbf{100.00} & \textbf{100.00} & \textbf{100.00} \\
005\_tomato\_soup\_can & 94.87 & \textbf{95.31} & 94.87 & 95.09 \\
006\_mustard\_bottle & \textbf{100.00} & \textbf{100.00} & \textbf{100.00} & \textbf{100.00} \\
007\_tuna\_fish\_can & 99.67 & 99.67 & 93.33 & \textbf{99.67} \\
008\_pudding\_box & 98.67 & \textbf{100.00} & \textbf{100.00} & \textbf{100.00} \\
009\_gelatin\_box & \textbf{100.00} & \textbf{100.00} & \textbf{100.00} & \textbf{100.00} \\
010\_potted\_meat\_can & 77.78 & \textbf{81.78} & 77.33 & 81.33 \\
011\_banana & 96.00 & 87.33 & \textbf{100.00} & \textbf{100.00} \\
019\_pitcher\_base & 60.89 & \textbf{100.00} & \textbf{100.00} & \textbf{100.00} \\
021\_bleach\_cleanser & 90.00 & \textbf{94.67} & 92.33 & 90.67 \\
024\_bowl & 17.33 & 50.00 & 66.00 & \textbf{82.67} \\
025\_mug & 91.33 & \textbf{100.00} & 97.33 & 98.00 \\
035\_power\_drill & \textbf{100.00} & \textbf{100.00} & 97.67 & 99.00 \\
036\_wood\_block & 69.33 & 81.33 & \textbf{100.00} & 93.33 \\
037\_scissors & \textbf{98.67} & 93.33 & 46.67 & 42.67 \\
040\_large\_marker & \textbf{100.00} & \textbf{100.00} & 99.33 & 98.00 \\
051\_large\_clamp & \textbf{99.33} & \textbf{99.33} & 86.00 & 98.67 \\
052\_extra\_large\_clamp & 78.67 & 75.33 & \textbf{99.33} & 91.33 \\
061\_foam\_brick & 96.00 & 94.67 & \textbf{100.00} & \textbf{100.00} \\
\hline
Mean & 88.98 & 92.49 & 90.98 & \textbf{92.56} \\
\bottomrule

\end{tabular}
\label{tab:ycbv}
\end{table}

\textbf{YCB-V.} 
On YCB-V we increase the $AR_{BOP}$ metric from 0.816 to 0.844 and the ADD-S 0.1 from 90.98 to 92.56 as listed in Table \ref{tab:ycbv}. Again, focusing on the BOP metrics, out of 8 symmetric objects, we beat our baseline on 4, not supporting the notion, that we learn a better understanding of symmetry. Instead we outperform on most of the texture-rich as well as texture-less objects, suggesting that EPRO-GDR can use, but does not rely on strong textures. 

\textbf{ITODD.} ITODD consists of industrial, metallic shapes and is the most difficult out of the 7 core datasets. We achieve an improvement of the metric result from 0.356 to 0.412. Since the ground truth of ITODD is not available we cannot compare object-level performance and can only report the 3 individual metrics, MSPD, MSSD, and VSD, all of which seem to improve to a similar degree, with MSPD benefiting slightly more than the other two. Since MSPD only compares visible discrepancies it is considered the most important for XR applications. A high score on this metric indicates a good fit. The great improvement on ITODD indicate that EPRO-GDR can deal better with metallic industrial shapes. \\
\textbf{T-Less.} Our method has some problems with the T-Less dataset. We still tend to achieve a high quality with many objects, outperforming GDRNPP on objects 10, 11, 21, and 26, for instance, but we face challenges with a few individual objects, reducing our average score. The offending objects are 14, 16, 27, 28, and 30. Please consider Table \ref{tab:quantitative3} for detailed, per object results. Template-based PFA also shows reduced performance with these texture-less objects, while SurfEmb still achieves high scores. 

\begin{table}[h]
\centering
\caption{Results with per object results on T-LESS. $AR_{BOP}$ metric, higher is better. Symmetric objects in \textit{italics}. The best result between GDRNPP and EPRO-GDR in \textbf{bold font}. }
\begin{tabular}{ccl>{\columncolor[gray]{0.9}}c>{\columncolor[gray]{0.9}}c}
\toprule
\multirow{2}{*}{} & \multicolumn{1}{c}{SurfEmb \cite{bop2021surfemb}} & \multicolumn{1}{c}{PFA \cite{PFA-BOP}} & \multicolumn{1}{c}{GDRNPP \cite{gdrnpp_bop}} & \multicolumn{1}{c}{Ours} \\
\midrule
\textit{Obj. 1} & 0.726 & 0.656 & \textbf{0.837} & 0.766 \\
\textit{Obj. 2} & 0.668 & 0.704 & \textbf{0.818} & 0.741 \\
\textit{Obj. 3} & 0.900 & 0.758 & \textbf{0.887} & 0.854 \\
\textit{Obj. 4} & 0.605 & 0.638 & \textbf{0.849} & 0.804 \\
\textit{Obj. 5} & 0.941 & 0.926 & 0.946 & \textbf{0.948} \\
\textit{Obj. 6} & 0.963 & 0.874 & 0.930 & \textbf{0.946} \\
\textit{Obj. 7} & 0.893 & 0.851 & 0.876 & \textbf{0.895} \\
\textit{Obj. 8} & 0.950 & 0.901 & \textbf{0.903} & 0.900 \\
\textit{Obj. 9} & 0.950 & 0.927 & 0.905 & \textbf{0.906} \\
\textit{Obj. 10} & 0.934 & 0.901 & 0.894 & \textbf{0.925} \\
\textit{Obj. 11} & 0.912 & 0.828 & 0.865 & \textbf{0.924} \\
\textit{Obj. 12} & 0.921 & 0.855 & 0.878 & \textbf{0.910} \\
\textit{Obj. 13} & 0.817 & 0.647 & \textbf{0.855} & 0.854 \\
\textit{Obj. 14} & 0.812 & 0.810 & \textbf{0.865} & 0.183 \\
\textit{Obj. 15} & 0.769 & 0.770 & \textbf{0.838} & 0.558 \\
\textit{Obj. 16} & 0.894 & 0.821 & \textbf{0.920} & 0.448 \\
\textit{Obj. 17} & 0.956 & 0.935 & \textbf{0.939} & 0.927 \\
\textit{Obj. 18} & 0.918 & 0.911 & 0.856 & \textbf{0.914} \\
\textit{Obj. 19} & 0.863 & 0.772 & \textbf{0.807} & 0.698 \\
\textit{Obj. 20} & 0.815 & 0.764 & \textbf{0.777} & 0.653 \\
\textit{Obj. 21} & 0.798 & 0.733 & 0.792 & \textbf{0.820} \\
\textit{Obj. 22} & 0.749 & 0.719 & 0.683 & \textbf{0.758} \\
\textit{Obj. 23} & 0.930 & 0.874 & 0.857 & \textbf{0.867} \\
\textit{Obj. 24} & 0.948 & 0.784 & 0.886 & \textbf{0.905} \\
\textit{Obj. 25} & 0.831 & 0.783 & 0.823 & \textbf{0.924} \\
\textit{Obj. 26} & 0.937 & 0.822 & 0.849 & \textbf{0.947} \\
\textit{Obj. 27} & 0.906 & 0.842 & \textbf{0.843} & 0.125 \\
\textit{Obj. 28} & 0.929 & 0.841 & \textbf{0.848} & 0.417 \\
\textit{Obj. 29} & 0.948 & 0.862 & \textbf{0.936} & 0.929 \\
\textit{Obj. 30} & 0.794 & 0.765 & \textbf{0.858} & 0.195 \\
\hline
Mean & 0.828 & 0.779 & \textbf{0.855} & 0.765 \\
\bottomrule
\end{tabular}
\label{tab:quantitative3}
\end{table}

\begin{table*}[h]
    \centering
    \begin{tabular}{|c|c|c|c|}
        \hline
          & \textbf{Sample 1} & \textbf{Sample 2} & \textbf{Sample 3} \\
        \hline
        \begin{sideways}RGB input\end{sideways} & \includegraphics[width=2.8cm]{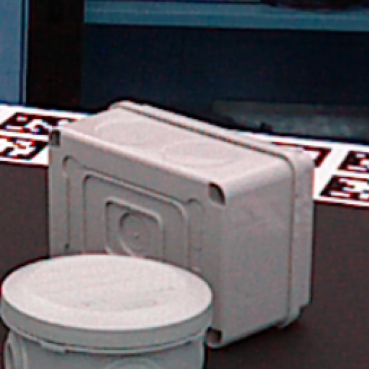} & \includegraphics[width=2.8cm]{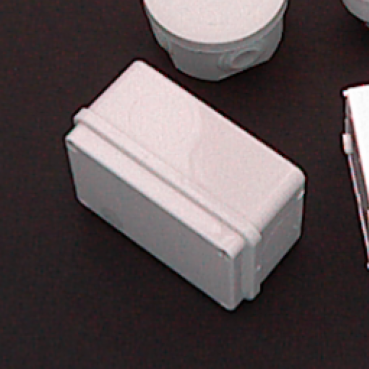} & \includegraphics[width=2.8cm]{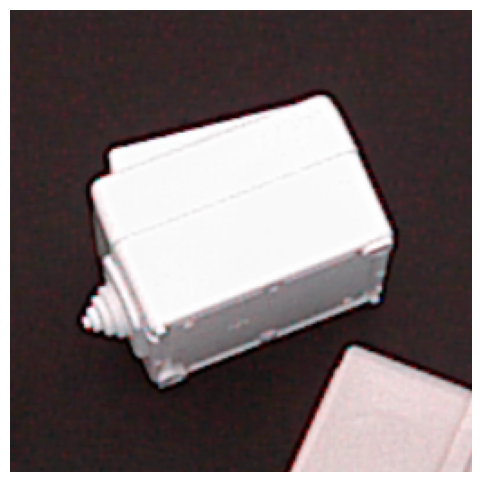} \\
        \hline
        \begin{sideways}predicted points\end{sideways} & \includegraphics[width=2.8cm]{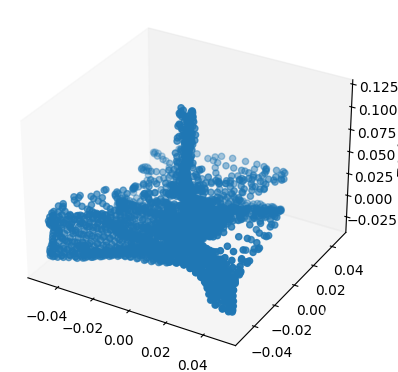} & \includegraphics[width=2.8cm]{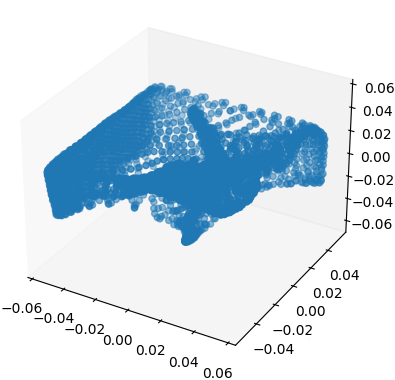} & \includegraphics[width=2.8cm]{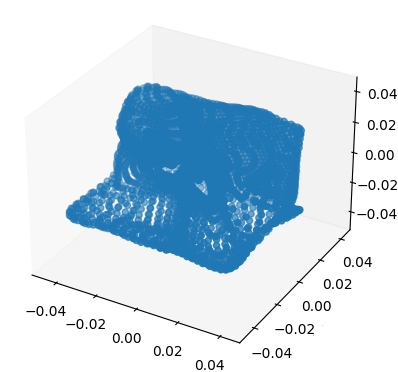} \\
        \hline
        \begin{sideways}GT points\end{sideways} & \includegraphics[width=2.8cm]{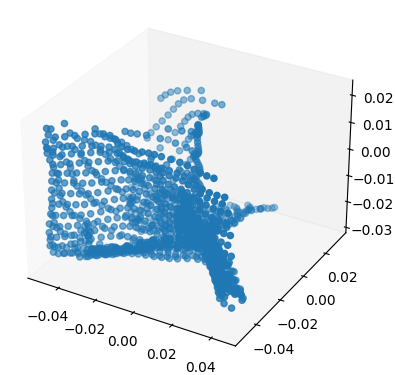} & \includegraphics[width=2.8cm]{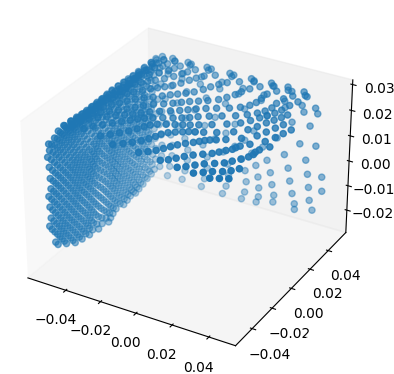} & \includegraphics[width=2.8cm]{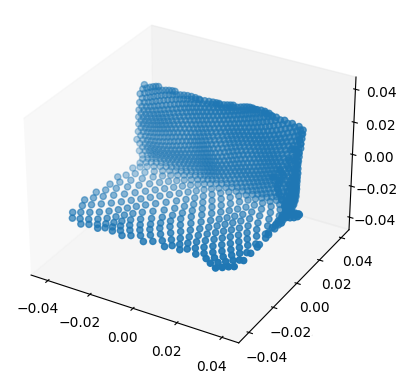} \\
        \hline
    \end{tabular}
    \caption{Single model training for T-Less object 27 (first two samples). Given an input sample, we show the 3D points as predicted by the model versus the ground truth. We find that the model has problems estimating the correct shape. For comparison we present object 26 (trained in a multiple objects, single model case) in column three. The shape is well understood as already indicated by the very high score of 0.947 for EPRO-GDR compared to only 0.849 for GDNRPP. }
    \label{fig:t-less_analysis}
\end{table*}

\subsection{Discussion}

EPRO-GDR outperforms GDRNPP on 3 out of 4 datasets. To indentify the challenges lie with T-Less, we delve deeper to find the source of our problem: Though there are some \href{https://github.com/ylabbe/cosypose/issues/7#issuecomment-689017593}{misalignment issues} between depth and RGB sensors in the T-Less dataset, 
we don't think this is the sole reason for the poor results on some objects. For further investigation, we pick out object 27, which has its score drop from 0.843 with GDRNPP to a mere 0.125 using EPRO-GDR, and we train a model solely on this single object. Since GDRNPP tends to perform significantly better in the single model per single object use case, we anticipated achieving much higher performance. While we did see improvements, we still did not reach the performance levels of GDRNPP. Our model tops out at 0.615, suggesting some deeper problem with specific geometric features. We show 3 samples with predicted 3D points and ground truth in Table \ref{fig:t-less_analysis}. For the poorly performing object 27, we note a discrepancy between the predicted 3D points at the bottom of the object and the corresponding ground truth. Specifically, the object is not as concave as the predicted points suggest. This inconsistency affects the calculation of re-projection error and the learning of the overall shape. On a closer look at the dataset, one reason for this is that there is not enough training data showing the bottom of the object. This issue may be less severe in single-object training but becomes more serious in multi-object training (as proved by the much improved performance of 0.615 vs. 0.125). Another possible reason is that the structure on the bottom makes it challenging to accurately determine the 3D points from the RGB information, as they create the illusion of a more concave bottom than is actually present. Indeed, the disorganized nature of the predicted 3D points suggests that the model does not fully comprehend the entire shape of the object. For comparison, we include a sample from the extremely well performing object 26, on which we drastically outperform GDRNPP, in column 3. Our method is capable of learning even the challenging, under-represented details of the texture-less T-Less objects, as demonstrated by our single-model training. To address outliers, we suggest increasing the number of training samples for problematic object views, giving them a higher weight during optimization. \\
All in all, EPRO-GDR outperforms its baseline on most objects, with T-Less containing a few very bad outliers. That being said, even on T-Less EPRO-GDR outperforms GDRNPP on 14 out of 30 objects making EPRO-GDR the overall better performing algorithm and placing it in line with the succession of incremental improvements starting with CDPN, GDR-Net, EPro-P$n$P, and GDRNPP. Considering the suitability for scene-level optimization, we deem EPRO-GDR to stand out among these choices, not only because of improved single-view accuracy, but by bringing the virtues of EPro-P$n$P's probability-based pose prediction to GDRNPP's design and by changing the prediction output from discrete poses to a probability distribution. The formulation as a pdf allows to sample multiple meaningful poses, the benefit of which is easy to see once one considers the often occurring possibility of visually non-distinguishable views. Also, typically ML algorithms are designed to also predict a score, representing how confident they are in their prediction. The probability of a pose sample is a meaningful representation for confidence. Future work might look into directly incorporating the pdfs provided by EPRO-GDR for scene-level optimization.


\section{Conclusion}
In this paper, we argue that a probabilistic formulation of the single-view 6D pose estimation problem is useful, because it allows to sample multiple likely pose candidates, and we show that it can improve the accuracy of pose estimates. To this end, we reformulated the state-of-the-art GDRNPP algorithm into our novel EPRO-GDR algorithm, which estimates a probability density distribution of poses. EPRO-GDR's ability to sample multiple pose estimates with a meaningful measure of uncertainty is a notable advancement over traditional single-view estimators for use cases that can incorporate the information of multiple estimates, such as scene-level optimization. Even when sampling just a single pose candidate (i.e. the mode of the distribution), EPRO-GDR shows superior quantitative results in pose estimation across three widely-used reference datasets. 

\bibliographystyle{IEEEtran.bst}
\bibliography{bib.bib}   

\end{document}